\documentclass{ifacconf}
\usepackage{algpseudocode}
\usepackage{amsmath}
\usepackage{amssymb}
\usepackage{graphicx}      
\usepackage{natbib}        
\renewcommand{\thefootnote}{}
\begin{document}
\begin{frontmatter}

\title{A Dual Quaternion based RRT* Path Planning Approach for Satellite Rendezvous and Docking} 

\author[First]{A. Stankovic} 
\author[Second]{M.K. Ben-Larbi} 
\author[First]{W.H. Müller}

\address[First]{TU Berlin, Chair of Continuum Mechanics and Consitutive Theory, 10623 Berlin , Germany (e-mail: a.stankovic@tu-berlin.de).}
\address[Second]{University of Würzburg, Chair of Space Informatics and Satellite Systems, 97070 Würzburg, Germany (e-mail: khalil.ben-larbi@uni-wuerzburg.de)}

\begin{abstract}                
This paper proposes a sampling-based motion planner that employs a dual quaternion representation to generate smooth, collision-free six-degree-of-freedom pose trajectories for satellite rendezvous and docking under keep-out zone constraints. The proposed planner integrates the dual quaternion algebra directly into an RRT* framework, thereby enabling natural screw motion interpolation in $SE(3)$. The dual quaternion-based RRT* has been implemented in Python and demonstrated on a representative multi-obstacle scenario. A comparison with a standard RRT* using separate translation and quaternion steering highlights the enhanced pose continuity and obstacle avoidance of the proposed method.
The present approach is purely kinematic in nature and does not take into account relative orbital dynamics. Consequently, the resulting path provides a preliminary estimate for a subsequent optimisation-based trajectory planner, which will refine the motion with dynamic constraints for the purpose of practical satellite rendezvous and docking missions.
\end{abstract}


\end{frontmatter}
\footnote{© 2025 the authors. This work has been accepted to IFAC for publication under a Creative Commons Licence CC-BY-NC-ND}
\section{Introduction}
Autonomous rendezvous and docking (RVD) is a key enabling technology for satellite servicing, debris removal, and future on-orbit assembly. This task requires precise planning of the six-degree-of-freedom (6-DOF) relative motion between the chaser and target spacecraft. Sampling-based motion planners such as Rapidly Exploring Random Trees (RRT) and their optimal variant RRT* are widely used for high-dimensional path planning. However, conventional implementations represent translation and rotation separately, which can lead to discontinuities and suboptimal interpolation in the $SE(3)$ group. To address this, an RRT* planner is proposed that uses a unit dual-quaternion representation to unify pose and enable screw motion interpolation. This work focuses on demonstrating smooth, collision-free path generation under a keep-out zone constraint.

\section{Related Work}

Sampling-based planners such as Rapidly Exploring Random Trees (RRT) and their optimal variant RRT* are widely used for motion planning in high-dimensional configuration spaces \citep{Karaman2011}. These planners are valued for their global exploration capability: by incrementally growing a tree that probabilistically covers the entire configuration space, they can find collision-free paths in complex, cluttered environments where local or gradient-based methods would fail. Recent surveys highlight the diversity of RRT* variants and their application domains \citep{Dey2022}, underlining their role as a foundation for many practical planning systems.

A central difficulty in applying such planners to rigid-body motion lies in the representation and interpolation of poses in $SE(3)$. The Open Motion Planning Library (OMPL) \citep{Sucan2012}, for instance, models translation in $\mathbb{R}^3$ and orientation in $SO(3)$ using quaternions. While straightforward to implement, this design typically relies on distance metrics that combine Euclidean and quaternion distances in an ad hoc fashion, which can introduce discontinuities and degrade path smoothness. To address these limitations, alternative formulations operate directly on the Lie group $SE(3)$, thereby ensuring mathematical consistency of interpolation and distance computation \citep{Park1995, Hauser2011}.

Among such representations, dual quaternions have been extensively studied in the robotics community as a compact algebraic tool for rigid-body motions \citep{Kenwright2012a, Kenwright2012b}. By unifying translation and rotation within a single framework, they naturally enable screw interpolation and continuity of pose trajectories. This has motivated their use in applications such as manipulator motion planning and local trajectory smoothing. For example, M{\"u}ller et al. \citep{Elbanhawi2014} employed dual quaternions in a sample-based planner for robot movement and tracking control. However, prior work has largely restricted their use to local refinements or control, rather than integration into asymptotically optimal, sampling-based global planners.

In parallel, optimization-based approaches have offered an alternative perspective on motion planning. Richards et al. \citep{Richards2002} formulated spacecraft rendezvous as a mixed-integer programming problem, achieving strict enforcement of constraints at the expense of high computational cost. Similarly, Kuwata et al. \citep{Kuwata2009} proposed a receding-horizon model predictive control scheme for UAVs that guarantees safety through closed-loop prediction. Nevertheless, these methods are computationally intensive and typically require good initializations. As a result, they are often combined with RRT*-style planners that provide globally feasible paths, which are subsequently refined into dynamically feasible trajectories through optimization.

Despite their theoretical appeal, dual quaternion-based approaches still face challenges when integrated into sampling-based planning. First, the definition of a distance metric must appropriately balance translational and rotational components, as poor weighting choices can distort exploration. Second, screw-path discretization must be sufficiently fine to ensure collision safety, which increases computational cost. Finally, planners based solely on kinematics neglect dynamics, control authority, and time-parametrization, preventing direct execution on real spacecraft without additional post-processing.

This paper addresses these gaps by proposing an RRT* formulation that directly integrates dual quaternion algebra into all stages of the planner: pose representation, sampling, steering, distance computation, and collision checking. The resulting method generates smooth, collision-free screw-motion trajectories in $SE(3)$, making it particularly well suited for satellite rendezvous and docking, where strict clearance with keep-out zones and continuity of six-degree-of-freedom motion are critical. The planner is demonstrated in representative scenarios and is intended as a high-quality kinematic path generator that can serve as an initial guess for subsequent dynamics-aware optimization.

\section{Mathematical Preliminaries}
This section offers a simplified introduction to the fundamentals of the Dual Quaternion Representation and the RRT* code in general, in addition to the theoretical interconnection of these concepts. A more comprehensive overview of the fundamental principles under discussion can be found in the work of~\cite{Kenwright2012b} and~\cite{lavalle2006planning}. 

\subsection{Dual Quaternion Representation}
A rigid-body pose in three-dimensional space is an element of the special Euclidean group $SE(3)$, defined as the set of all rigid transformations combining rotation and translation:
\begin{equation}
SE(3) = \{ (\mathbf{R},\, \vec{t}) \mid \mathbf{R} \in SO(3),\; \vec{t} \in \mathbb{R}^3 \}.
\end{equation}
While $SE(3)$ is commonly represented by a $4 \times 4$ homogeneous transformation matrix, this work uses the dual quaternion algebra to encode the same rigid-body transformation more compactly.

A unit dual quaternion is defined as
\begin{equation}
    \hat{q} = \boldsymbol{q}_r + \epsilon \boldsymbol{q}_d, 
    \qquad \boldsymbol{q}_d = \tfrac{1}{2} \boldsymbol{p}\, \boldsymbol{q}_r, 
    \qquad \boldsymbol{p} = [0,\, \vec{t}], \label{eq:unit_dual_quat}
\end{equation}
where $\boldsymbol{q}_r$ is a unit quaternion representing rotation, $\vec{t} \in \mathbb{R}^3$ is a translation vector, and $\epsilon$ is the dual unit with the property $\epsilon^2 = 0$. The dual part $\boldsymbol{q}_d$ encodes the translation relative to the rotation $\boldsymbol{q}_r$. Rigid-body transformations can be composed by dual quaternion multiplication.

\subsection{Screw motion}
\begin{figure}
    \centering
    \includegraphics[width=0.8\linewidth]{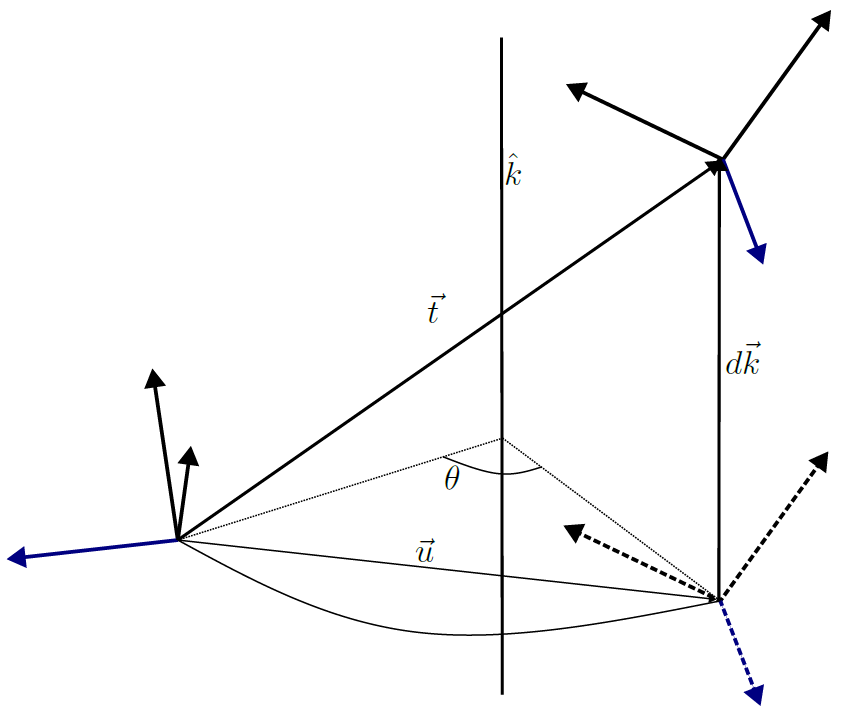}
    \caption{The displacement $\vec{t}$ shown as the sum of the translation $\vec{u}$ generated by the rotation $\theta$ about the screw axis $\hat{\boldsymbol{k}}$, and the translation $d\vec{k}$ along the the screw axis}
    \label{fig:screw_motion}
\end{figure}

To give a better idea of how dual quaternions operate, these are explained, as well as the underlying interpolation using screw theory. This states that according to Chasles’ theorem, any rigid-body displacement can be represented as a rotation $\theta$ about an axis combined with a translation $d$ along the same axis, as depicted in Fig. \ref{fig:screw_motion}.

This so-called screw motion can be compactly expressed in dual quaternion form as
\begin{equation}
    \hat{q} = \cos \tfrac{\hat{\theta}}{2} + \hat{{k}}\, \sin \tfrac{\hat{\theta}}{2},
    \label{eq:dualquat_screw}
\end{equation}
where $\hat{k}$ is the unit dual vector of the screw axis and 
\begin{equation}
    \hat{\theta} = \theta + \epsilon d
\end{equation}
is the dual angle \citep{Sjoberg2017,McCarthy1990}. This generalizes the familiar axis–angle form of quaternions: the real part encodes pure rotation, while the dual part captures translation along the same axis. Importantly, for screw motions the order of rotation and translation commutes, unlike in general $SE(3)$ compositions.

\subsection{Interpolation with ScLERP}
Screw motions provide a natural foundation for interpolation between poses. The screw linear interpolation (\textsc{ScLERP}) between two dual quaternions $\hat{q}_1$ and $\hat{q}_2$ is defined as
\begin{equation}
    \hat{q}(s) = \hat{q}_1 \, \big( \hat{q}_1^{-1} \hat{q}_2 \big)^s, 
    \qquad s \in [0,1].
    \label{eq:ScLERP}
\end{equation}
Geometrically, \textsc{ScLERP} traces the shortest screw motion connecting $\hat{q}_1$ and $\hat{q}_2$, producing a continuous helical trajectory in $SE(3)$. This contrasts with conventional approaches that interpolate translation and rotation separately, which can result in discontinuities or misalignment between motion components \citep{Kenwright2012b}.

\subsection{Rigid-Body Kinematics in Dual Quaternion Form}

Rigid-body motion in the six-degree-of-freedom configuration space $SE(3)$ is governed by purely kinematic equations that describe how pose evolves under a twist. Using the dual quaternion representation, the pose of a rigid body is denoted by a unit dual quaternion as stated in Equation~\ref{eq:unit_dual_quat}.

The rigid-body kinematics can then be expressed as
\begin{equation}
    \dot{\hat{q}}(t) = \tfrac{1}{2}\,\hat{\xi}(t)\,\hat{q}(t),
    \label{eq:dualquat_kinematics}
\end{equation}
where $\hat{\xi}(t)$ is the \emph{dual quaternion twist} defined as
\begin{equation}
    \hat{\xi}(t) = \boldsymbol{\omega}(t) + \epsilon \boldsymbol{v}(t),
\end{equation}
with $\vec{\omega} \in \mathbb{R}^3$ the angular velocity and $\vec{v} \in \mathbb{R}^3$ the linear velocity expanded to quaternions.

Equation~\eqref{eq:dualquat_kinematics} is the dual quaternion equivalent of the standard homogeneous transformation kinematics
\begin{equation}
    \dot{T}(t) = 
    \begin{bmatrix}
        \vec{\omega}_\times & \vec{v} \\
        0 & 0
    \end{bmatrix} T(t),
    \label{eq:homog_kinematics}
\end{equation}
where $\omega_\times$ is the skew-symmetric angular velocity matrix.

In the present work, the planner does not numerically integrate~\eqref{eq:dualquat_kinematics} during sampling. Instead, local steering between two poses is performed using screw linear interpolation (ScLERP) as in Equation \ref{eq:ScLERP}, which represents a discrete realization of the continuous rigid-body kinematics. 


\subsection{Problem formulation}
The planning problem describes the six-degree-of-freedom pose of the chaser relative to the target. Each pose is represented as a unit dual quaternion $\hat{q}$.

Given an initial pose $\hat{q}_{\text{init}}$ and a desired goal pose $\hat{q}_{\text{goal}}$, the objective is to generate a continuous sequence of feasible poses:
\begin{align}
\mathcal{P} = \{\hat{q}_0, \hat{q}_1, \ldots, \hat{q}_N\}
\quad \text{with} \quad
\hat{q}_0 = \hat{q}_{\text{init}}, \;
\hat{q}_N = \hat{q}_{\text{goal}},
\end{align}
that connect the initial pose to the goal while avoiding collision with defined keep-out zones.

The configuration space is bounded within the prescribed limits on translation and full rotation in $SO(3)$. Obstacles are modelled as fixed spherical keep-out regions $\mathcal{O} = \{O_1, O_2, \ldots, O_K\}$, 
each defined by a center position $\vec{c}_k$ and radius $ r_k$. A candidate path $\mathcal{P}$ is valid if every intermediate pose maintains the required clearance:
\begin{align}
|\vec{t}_i - \vec{c}_k\| > r_k, 
\quad \forall O_k \in \mathcal{O}, \; \forall i \in [0, N],
\end{align}

where $\mathbf{t}_i$ is the translation extracted from $\hat{q}_i$.

This solution is used as a global motion planner that probabilistically explores the entire configuration space to find a feasible, collision-free path.
The planner does not enforce relative dynamics or thrust constraints at this stage; instead, the resulting collision-free pose trajectory serves as an initial guess for a future optimization-based trajectory planner that refines the motion with dynamics, control limits, and performance objectives. 

\subsection{General RRT* algorithm}

The RRT* algorithm is a sampling-based motion planner that incrementally builds a tree in the configuration space by randomly sampling feasible states and connecting them with a local steering function~\cite{Karaman2011}. Compared to the original RRT, RRT* includes a rewiring step that incrementally improves the cost of the tree by reconnecting new nodes through shorter or lower-cost paths, ensuring asymptotic optimality.

At each iteration, a random state $x_{\text{rand}}$ is sampled, the nearest node $x_{\text{near}}$ in the tree is found, and a new node $x_{\text{new}}$ is generated by steering toward the sample. If the edge between $x_{\text{near}}$ and $x_{\text{new}}$ is collision-free, the new node is added and the local neighborhood is rewired to minimize the total cost. The resulting tree approximates an optimal solution as the number of samples grows to infinity.

\begin{algorithm}[h]
\caption{General RRT* Algorithm}
\begin{algorithmic}[1]
\State \textbf{Input:} Initial state $x_{\text{init}}$, goal region $\mathcal{X}_{\text{goal}}$, obstacle space $\mathcal{X}_{\text{obs}}$
\State \textbf{Initialize} tree with root node $x_{\text{init}}$
\While{goal not reached}
    \State $x_{\text{rand}} \leftarrow$ \textsc{SampleRandomState}()
    \State $x_{\text{near}} \leftarrow$ \textsc{NearestNeighbor}($x_{\text{rand}}$)
    \State $x_{\text{new}} \leftarrow$ \textsc{Steer}($x_{\text{near}},\, x_{\text{rand}}$)
    \If{\textsc{CollisionFree}($x_{\text{near}},\, x_{\text{new}},\, \mathcal{X}_{\text{obs}}$)}
        \State Add $x_{\text{new}}$ to tree
        \State \textsc{Rewire}(tree, $x_{\text{new}}$)
    \EndIf
\EndWhile
\State \textbf{Output:} Feasible path from $x_{\text{init}}$ to $\mathcal{X}_{\text{goal}}$
\end{algorithmic}
\label{alg:rrtstar_general}
\end{algorithm}
\noindent

While the general RRT* framework can plan in any configuration space given a valid state representation, distance metric, and steering function, it typically handles translation and rotation separately when applied in $SE(3)$.

\subsection{Dual Quaternion-Based RRT* Algorithm}

In the proposed RRT* planner, each tree node represents a pose in $SE(3)$ using a unit dual quaternion. The distance metric for nearest-neighbour search combines translational and rotational components:
\begin{equation}
d(\hat{q}_1,\, \hat{q}_2) = w_t\, \| \vec{t}_1 - \vec{t}_2 \| + w_r\, \cos^{-1}(| \boldsymbol{q}_{r1} \cdot \boldsymbol{q}_{r2} |),
\end{equation}
where $\vec{t}_1, \vec{t}_2$ are the extracted translations and $\boldsymbol{q}_{r1}, \boldsymbol{q}_{r2}$ are the corresponding rotation quaternions.The distance metric combines Euclidean translation and geodesic quaternion distance, following standard practice for pose metrics in $SE(3)$~\cite{Kenwright2012a}.

This representation preserves the full $SE(3)$ structure while enabling efficient sampling, nearest-neighbor search, and screw interpolation inside the planner~\cite{Kenwright2012a}.

The proposed motion planner uses an asymptotically optimal RRT* framework to solve the kinematic pose planning problem formulated above. Each node in the search tree encodes a pose as a unit dual quaternion $\hat{q}$. 

The steering function uses screw linear interpolation (\textsc{Sclerp}) to generate an incremental motion from the nearest node toward the random sample adapted from equation \ref{eq:ScLERP}:
\begin{equation*}
\hat{q}_{\text{new}} = \hat{q}_{\text{near}} 
\Big( \hat{q}_{\text{near}}^{-1} \hat{q}_{\text{rand}} \Big)^\alpha,
\quad \alpha \in (0,1).
\end{equation*}

Each candidate edge is discretized along the screw path. At each intermediate pose, the planner checks for violations of all keep-out zone constraints. If the edge is collision-free, the new node is added to the tree, and rewiring is performed to maintain asymptotic optimality.

The output is a smooth, piecewise screw motion trajectory that connects the initial pose to the goal pose while avoiding all obstacles. Following the structure summarized by \cite{Dey2022}, the proposed dual quaternion-based RRT* follows the standard expansion, steering, and rewiring cycle as \ref{alg:rrtstar_general}.

The steering function used in the dual quaternion RRT* is implemented through screw linear interpolation (ScLERP) in equation~\eqref{eq:ScLERP}. 
This operation provides a discrete realization of the continuous rigid-body kinematics described by equation~\eqref{eq:dualquat_kinematics}. 
Rather than numerically integrating the dual quaternion differential equation, the planner advances in discrete increments along a screw motion connecting two sampled poses. 
Thus, each edge expansion follows a kinematically consistent trajectory in $SE(3)$ while maintaining computational efficiency. 

The standard tree expansion, nearest-neighbor search, and rewiring logic are preserved to maintain the asymptotic optimality of RRT* as shown in Algorithm~\ref{alg:rrtstar_general}.

\section{Simulation Setup}

For this study the main objective is to adapt an open-source implementation of the RRT* algorithm~\cite{rrtalgorithmsgithub} to incorporate full pose planning by adding a rotation component to each node, following the example of the Open Motion Planning Library (OMPL)~\cite{Sucan2012}. In the original version orientation representation is done via standard quaterions.

\subsection{Test scenario for Obstacle Avoidance}
\begin{figure}[h]
  \centering
  \includegraphics[width=\linewidth]{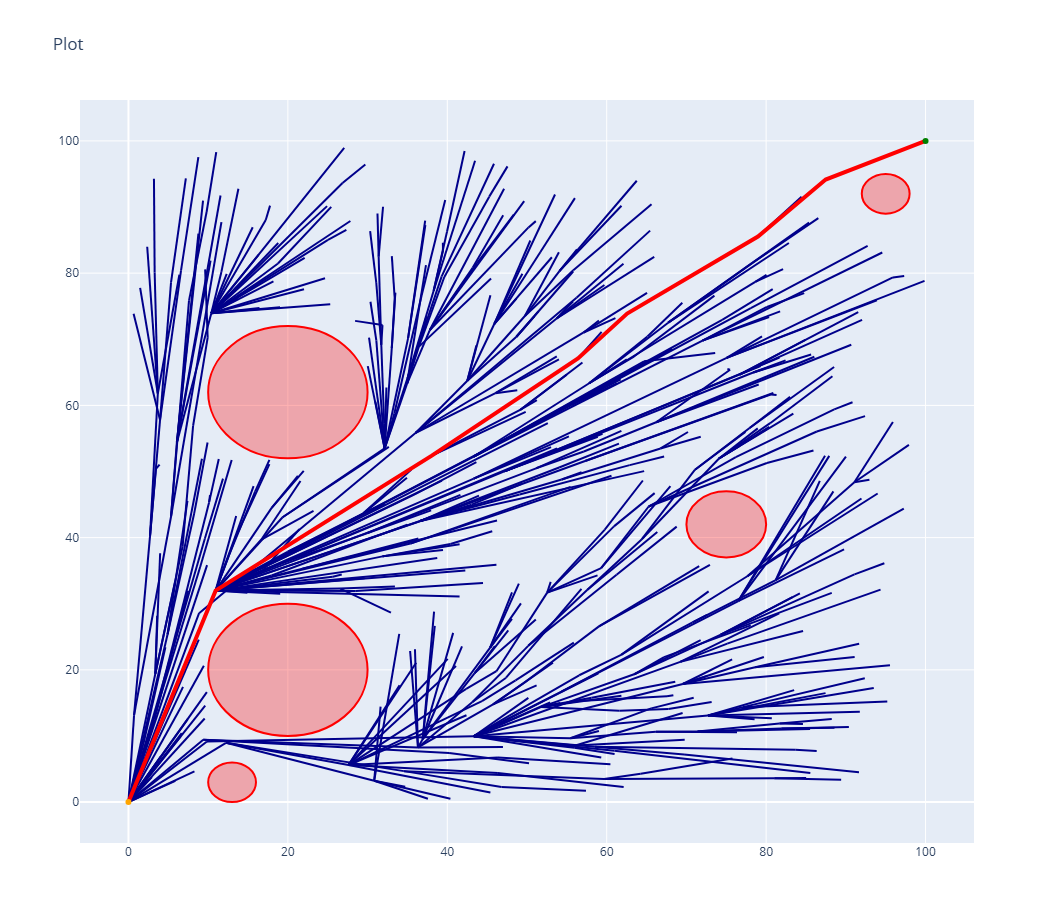}
  \caption{RRT* in 2D translation space with multiple circular keep-out zones shown to illustrate obstacle avoidance without explicit rotation representation}
  \label{fig:keepout_scenario}
\end{figure}

As a starting point, an open-source implementation of a basic translational RRT* planner in two dimensions was employed (see reference {rrtAlgorithms}). The reference code provides a minimal but functional framework for sampling, collision checking, and tree expansion. The algorithm was initially adapted to our scenario by adjusting the sampling domain and collision-checking routines. This is done to represent the desired obstacle and keep-out zones relevant for spacecraft proximity operations and ensure that the planner correctly avoided forbidden regions while still producing asymptotically optimal paths in the translational workspace. However, the implementation of proximity operations would normally necessitate the establishment of a single keep-out zone surrounding the target.
The scenario type is customary in the context of investigating RRT-algorithms and could serve as a basis for planning non-proximity path planning. A visualization of the modified 2D RRT* result is shown in Fig.~\ref{fig:keepout_scenario}.

\begin{figure}[h]
  \centering
  \includegraphics[width=\linewidth]{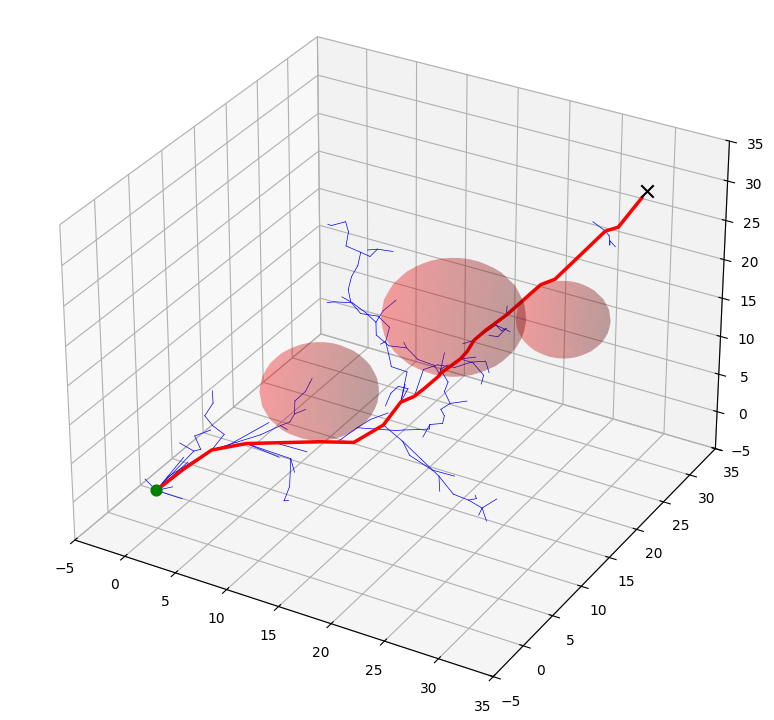}
  \caption{RRT* in 3D translation space with circular keep-out zones without explicit rotation representation}
  \label{fig:keepout_scenario_3D}
\end{figure}

The position and number of circular keep-out zones in the underlying code of Fig. \ref{fig:keepout_scenario} are randomly generated within the specified field, while the start and end positions remain. The functionality is verified based on the convergence of the algorithm.\\
Under the same premises, a 3D simulation of RRT* with spherical keep-out zones was also adapted, as shown in Fig.~\ref{fig:keepout_scenario_3D}, and validated on convergence by evaluation of the result.

\subsection{Baseline and Dual Quaternion Algorithm}

As a baseline, an RRT* planner derived from \cite{Sucan2012} is implemented and adjusted in $SE(3)$ where each node encodes a pose as $(x, y, z, q_x, q_y, q_z, q_w)$. Translation is represented in $\mathbb{R}^3$ and orientation by a unit quaternion. The local steering function advances linearly in position while the attitude is interpolated using quaternion spherical linear interpolation (SLERP). 
The distance metric combines Euclidean translation with the geodesic quaternion distance,
\begin{equation*}
    d(\vec{x}_1, \boldsymbol{q}_1; \vec{x}_2, \boldsymbol{q}_2) = \|\vec{x}_1 - \vec{x}_2\| + w \, \theta(\boldsymbol{q}_1, \boldsymbol{q}_2)
\end{equation*}
where $\theta(\boldsymbol{q}_1, \boldsymbol{q}_2)$ is the relative rotation angle. Candidate edges are discretized 
and collision-checked against spherical keep-out zones. While this approach generates 
feasible six-degree-of-freedom trajectories, it treats translation and rotation separately.

\begin{figure}[h]
  \centering
  \includegraphics[width=\linewidth]{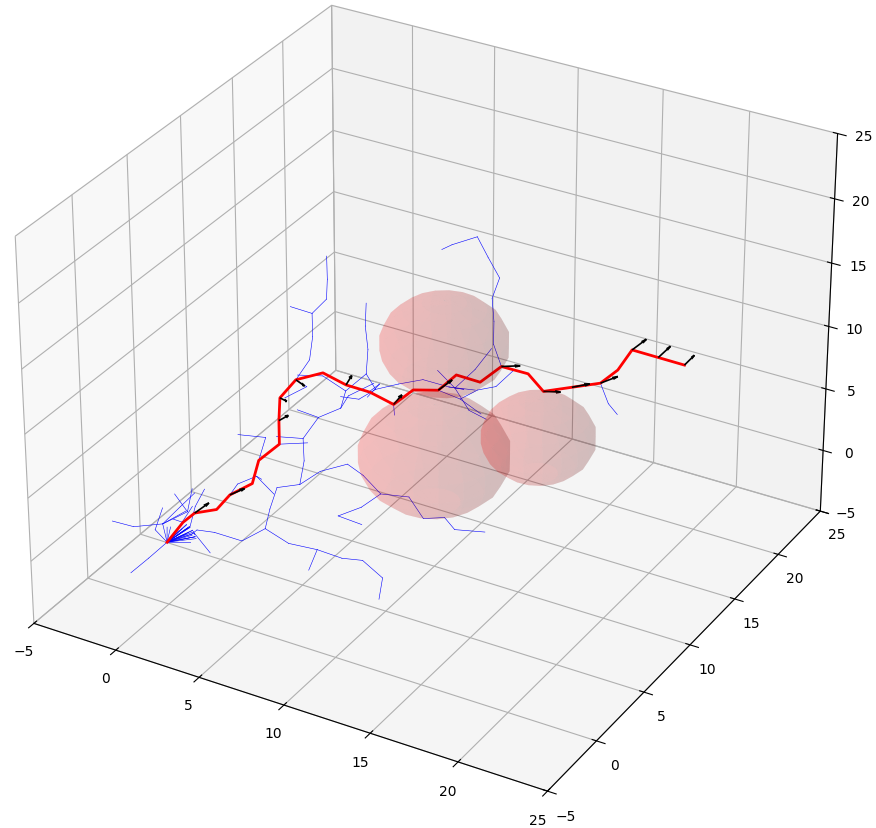}
  \caption{Baseline SE(3) RRT* path with spherical obstacles. Nodes and edges of the exploration tree are plotted in blue, and orientation arrows along the path indicate the rotation of a chosen body axis.}
  \label{fig:rrt_standard}
\end{figure}

\begin{figure}[h]
  \centering
  \includegraphics[width=\linewidth]{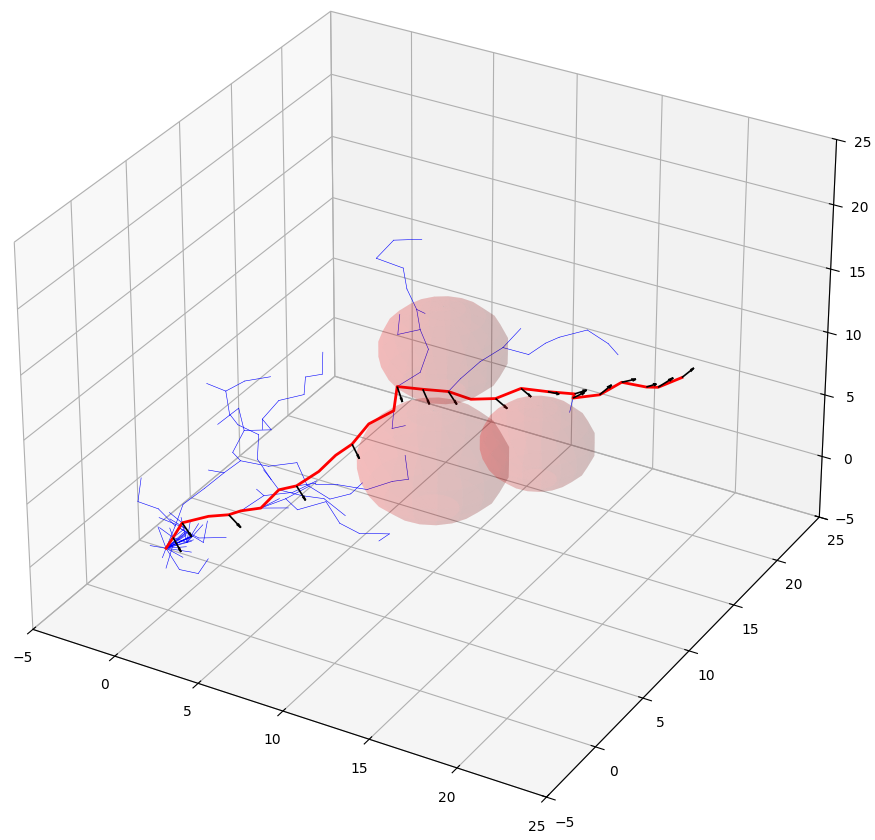}
  \caption{Dual-quaternion RRT* path, where ScLERP ensures smooth interpolation of translation and rotation with rotational evolution along the trajectory visible by the orientation arrows}
  \label{fig:rrt_dual}
\end{figure}

The comparison between the baseline SE(3) RRT* and the dual-quaternion (DQ) RRT* planner highlights several important differences in path representation, computational cost, and path smoothness.  

In order to provide a meaningful comparison between the two algorithms, the specifications for the end and start positions and the random seeding for the keep-out zones were set to be identical. In the specific scenarios depicted in Fig. \ref{fig:rrt_standard} and \ref{fig:rrt_dual}, the start orientation for both planners corresponds to an identity rotation (no initial rotation), while the goal orientation encodes a rotation of $\pi/2$ about the $z$-axis, achieved via a unit quaternion in SE(3) and a corresponding dual quaternion in the DQ representation. The plotted trajectories show that both planners successfully navigate around obstacles, but the DQ path maintains a more consistent rotational evolution along the trajectory, reflecting the benefits of the dual quaternion formulation in SE(3) motion planning.
In general, the DQ planner produces smoother paths in terms of combined translational and rotational motion, particularly visible in the orientation arrows along the path, while the baseline RRT* can exhibit slightly less coherent rotational transitions due to the separate treatment of translation and rotation. However, this improvement in smoothness comes at a higher computational cost: the DQ planner requires additional operations for dual quaternion logarithms, exponentials, and ScLERP, which can slow down the planning process.

\section{Results and Outlook}

The proposed dual quaternion-based RRT* algorithm successfully generated smooth, collision-free six-degree-of-freedom pose trajectories in the representative rendezvous scenario with multiple spherical keep-out zones. Compared to the initial implementation of RRT*, the dual quaternion formulation provides improved continuity of pose evolution. This is particularly visible in the rotational component: while the baseline planner interpolates \(\vec{t} \in \mathbb{R}^3\) linearly and uses \(\boldsymbol{q} \in \mathbb{H}\) for attitude via SLERP, the dual quaternion planner encodes the full pose as a unit dual quaternion \(\hat{q} \in \mathbb{DH}\) and advances along a screw motion via ScLERP. As a result, the final path avoids discontinuities between translation and rotation, generating a more physically meaningful trajectory.

The numerical experiments demonstrate that the dual quaternion-based planner can respect keep-out constraints while ensuring continuous screw motion in the presence of obstacles. However, these improvements come at the price of higher computational cost, as the ScLERP steering and dual quaternion algebra require additional logarithmic and exponential operations compared to the baseline method.

The present work is limited to purely kinematic path planning. The rigid-body kinematic equation \ref{eq:dualquat_kinematics}, is only realized in discrete form through screw interpolation. Dynamic feasibility, actuation limits, and temporal parametrization are not enforced at this stage. Consequently, the resulting path cannot be directly executed on a spacecraft without further processing.

Future work will therefore focus on embedding the proposed planner into a hierarchical architecture. The dual quaternion RRT* will serve as a fast global explorer in \(SE(3)\), generating an initial kinematic solution that satisfies geometric constraints. This path will then be refined by an optimization-based trajectory planner that incorporates orbital dynamics, control authority, and mission objectives such as fuel or time minimization. Such a combined framework is expected to bridge the efficiency of sampling-based planning with the accuracy of dynamics-aware trajectory optimization, ultimately enabling practical application to autonomous satellite rendezvous and docking.


\bibliography{ifacconf}             
                                                   








\appendix
\end{document}